\documentclass[conference]{IEEEtran}
 \IEEEoverridecommandlockouts 

\usepackage{cite}
\usepackage{amsmath,amssymb,amsfonts}
\usepackage{graphicx}
\usepackage{booktabs}
\usepackage{threeparttable}
\usepackage{array}
\usepackage{xcolor}

\begin{document}

\title{Task-Oriented Semantic Feature Transmission for Multi-Task Satellite Remote Sensing over Low-SNR Channels%
\thanks{Code: https://github.com/IntelliSensing/MTPjscc.git}
}

\author{
\IEEEauthorblockN{Shuoyuan Sun, Hongyu Wang, Mugen Peng, and Wenjia Xu*\thanks{*Corresponding author: Wenjia Xu.}}
\IEEEauthorblockA{
State Key Laboratory of Networking and Switching Technology\\
Beijing University of Posts and Telecommunications\\
Email: xuwenjia@bupt.edu.cn, shuoyuansun@bupt.edu.cn
}
}
\maketitle

\begin{abstract}

Conventional satellite remote sensing transmission follows a reconstruct-then-infer paradigm that optimizes pixel-level fidelity, creating an objective mismatch with downstream tasks such as classification and detection, especially at low SNR. This paper investigates a task-oriented framework that bypasses image reconstruction and directly transmits semantic features extracted by a multitask-pretrained backbone. A lightweight channel adaptation module (CAM) compresses feature dimensionality for bandwidth reduction, and a feature restorer recovers task-relevant structure after channel corruption. With the backbone frozen, the CAM and task-specific downstream heads are jointly optimized with task and feature-level supervision under random-SNR training. Under the adopted AWGN setting, experiments on scene classification and object detection show consistent gains over reconstruction-oriented JSCC baselines across different SNR conditions, with the largest improvements in the low-SNR regime.
\end{abstract}


\section{Introduction}

Low-Earth-orbit (LEO) satellite constellations provide global Earth observation and time-sensitive remote sensing, yet their downlinks face strict spectrum constraints and wide SNR variation driven by elevation angle, atmospheric attenuation, interference, and orbital geometry. Meanwhile, remote sensing is shifting from image delivery to immediate decision-making—scene classification, horizontal bounding-box (HBB) detection, and oriented bounding-box (OBB) detection—requiring the communication layer to serve downstream task utility rather than image recovery alone.

Existing semantic communication systems for remote sensing largely follow a reconstruct-then-infer paradigm. Deep joint source--channel coding (DeepJSCC)~\cite{bourtsoulatze2019deepjscc} methods such as SwinJSCC~\cite{yang2025swinjscc}, MambaJSCC~\cite{wu2026mambajscc}, and NTSCC~\cite{dai2022ntscc} map source images directly to channel symbols, but optimize pixel-level fidelity (MSE or SSIM). This creates a fundamental objective mismatch: the communication module preserves pixel-level details irrelevant to downstream classification or detection. At low SNR, reconstructed images suffer disproportionate task-metric drops because noise-induced artifacts disrupt the discriminative cues that downstream models rely on.

\begin{figure}
    \centering
    \includegraphics[width=1\linewidth]{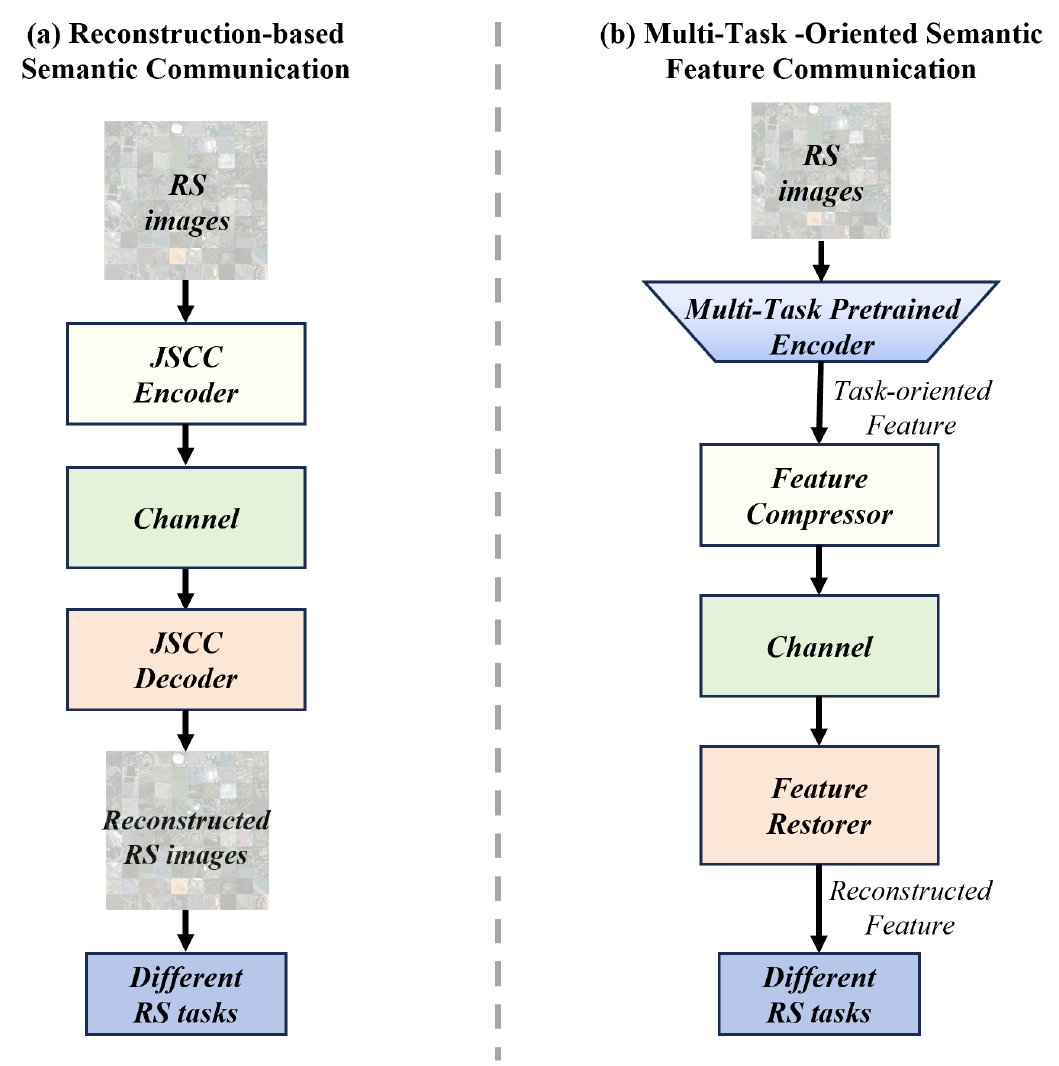}
    \caption{(a)~Conventional reconstruction-based image transmission versus (b)~proposed task-oriented semantic feature transmission. The proposed method extracts features with a multitask-pretrained backbone, transmits them through a CAM, and performs downstream tasks directly from recovered features without image reconstruction.}
    \label{fig:pipeline_comparison}
\end{figure}

Task-oriented semantic communication~\cite{bao2011semantic,xie2021deepsc} offers an alternative: transmitting only representations directly useful for downstream tasks—object-level structure and category-discriminative patterns—making more effective use of scarce downlink resources.

A key enabler is multitask-pretrained remote sensing backbones. Under the multi-task pretraining (MTP) paradigm~\cite{wang2024mtp}, a shared backbone trained jointly on semantic segmentation, instance segmentation, and rotated object detection yields transferable representations across downstream tasks. We adopt such a backbone as the semantic feature extractor, using its high-level features as the transmitted payload in place of the raw image.

However, noisy channels can degrade feature discriminative quality. We therefore introduce a channel adaptation module (CAM) between the pretrained backbone and downstream task heads. A feature compressor reduces channel dimensionality for bandwidth-constrained transmission, and a feature restorer recovers task-relevant structure after channel corruption. Combined with random-SNR training, the CAM preserves downstream task utility without reconstructing images.


This paper makes three main contributions. First, we propose a task-oriented semantic feature transmission framework for satellite remote sensing, where a multitask-pretrained backbone extracts transferable high-level representations and the communication pipeline transmits semantic features instead of reconstructed images. The framework is instantiated across scene classification, horizontal object detection, and oriented object detection to examine whether feature-domain transmission better aligns communication with downstream task utility. Second, we develop a lightweight channel adaptation module (CAM), composed of a feature compressor and a feature restorer, and train it jointly with each downstream task head under random-SNR conditions to improve robustness to channel distortion while maintaining transmission efficiency. Third, experiments on EuroSAT, DIOR, and DIOR-R show that, under the same compression ratios, backbone, task heads, AWGN evaluation protocol, and clarified separation between pretraining and downstream evaluation data, the proposed framework consistently outperforms reconstruction-oriented JSCC methods, with the most significant gains observed in the low-SNR regime.

\section{System Overview and Task-Oriented Feature Extraction}

The proposed framework consists of three stages: (1)~a frozen multitask-pretrained backbone extracts task-oriented semantic features from a remote sensing image; (2)~a channel adaptation module (CAM) compresses, transmits, and recovers these features through a noisy channel; and (3)~downstream task heads perform inference directly on the recovered features without image reconstruction. This section describes the feature extraction stage; the CAM is presented in Section~III.

Following MTP paradigm~\cite{wang2024mtp}, a shared backbone is trained with joint supervision from semantic segmentation, instance segmentation, and rotated object detection on the SAMRS ~\cite{wang2023samrs}. The resulting representations capture transferable task-relevant structure, including category-discriminative patterns, spatial relationships, and multi-scale object semantics, and generalize across downstream classification and detection applications. These multitask-pretrained features serve as the semantic payload for transmission.

We adopt a stage-wise pretraining scheme and denote the resulting backbone as IMP+MTP. IMP refers to the ImageNet-22K pretrained initialization of InternImage-XL~\cite{wang2023internimage}, and MTP denotes the subsequent supervised multi-task pretraining stage on SAMRS. The backbone therefore starts from a strong generic visual checkpoint and is further pretrained with remote sensing task supervision, reducing the gap between generic pretraining and downstream dense prediction.
\subsection{Backbone and Feature Pyramid Interface}
InternImage-XL~\cite{wang2023internimage} serves as the backbone. It employs dynamic sparse kernels and adaptive spatial aggregation for long-range context with locality and efficiency suited to dense visual tasks, and uses layer normalization, feed-forward networks, and GELU activations for stable optimization.

Given an input image $x \in \mathbb{R}^{3 \times H \times W}$, the encoder produces a hierarchy of multi-scale feature maps:
\begin{equation}
\{z_{\ell}\}_{\ell=1}^{L} = f_{\theta}(x), \qquad z_{\ell} \in \mathbb{R}^{C_{\ell} \times H_{\ell} \times W_{\ell}} .
\end{equation}
This hierarchy serves as a shared feature pyramid with spatial resolutions $1/4$, $1/8$, $1/16$, $1/32$ of the input, encouraging transferable rather than task-specific representations.

\subsection{Multi-Task Pretraining}
Three dense prediction heads—UperNet~\cite{xiao2018upernet} (semantic segmentation), Mask R-CNN~\cite{he2017maskrcnn} (instance segmentation), and Oriented R-CNN~\cite{xie2021orientedrcnn} (rotated detection)—are attached to the shared backbone and optimized jointly on SAMRS (dataset details in Section~IV).

Let $\mathcal{L}_{\text{rod}}$ denote the rotated detection loss, let $\mathcal{L}_{\text{ins}}^{b}$ and $\mathcal{L}_{\text{ins}}^{m}$ denote the instance box loss and instance mask loss, respectively, and let $\mathcal{L}_{\text{sem}}$ denote the semantic segmentation loss. Since SAMRS is composed of three subsets ($i \in \{1,2,3\}$), the total objective is:
\begin{equation}
\mathcal{L} = \sum_{i=1}^{3} \left( \mathcal{L}_{\text{rod}}^{i} + \mathcal{L}_{\text{ins}}^{i,b} + \mathcal{L}_{\text{ins}}^{i,m} + \mathcal{L}_{\text{sem}}^{i} \right) .
\end{equation}
These heads are not retained; their complementary supervision teaches cross-task semantic structure that single-task training cannot capture.

\begin{figure*}[t]
    \centering
    \includegraphics[width=\textwidth]{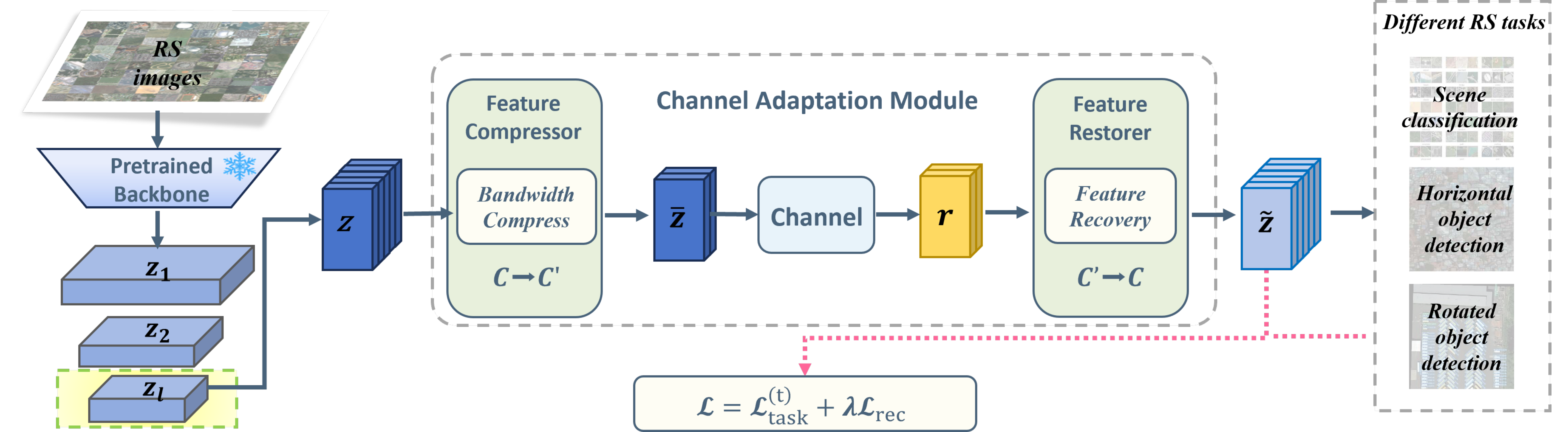}
    \caption{Overall architecture. A frozen multitask-pretrained backbone extracts task-oriented features; the last-stage feature map is compressed by the feature compressor ($1\!\times\!1$ Conv + GN), transmitted through the channel, and recovered by the feature restorer to obtain $\tilde{z}$. The recovered feature is fed directly into task heads: GAP + linear classifier for classification, or FPN + detection heads for HBB/OBB detection—no image is reconstructed.}
    \label{fig:overall_framework}
\end{figure*}

\section{Channel Adaptation Module for Task-Oriented Transmission}

This section details the CAM that transmits the payload $z$ over a noisy, bandwidth-constrained channel. It consists of a feature compressor (dimensionality reduction) and a feature restorer (dimensionality recovery after corruption). The entire pipeline operates in the feature domain with no image reconstruction.

\subsection{Payload Selection}
From the hierarchy $\{z_{\ell}\}_{\ell=1}^{L}$ produced by the frozen backbone, we select the last-stage feature map as the payload:
\begin{equation}
z \triangleq z_{L} \in \mathbb{R}^{C \times H' \times W'}.
\end{equation}
Shallower maps $\{z_\ell\}_{\ell=1}^{L-1}$ are not transmitted. Classification uses only $\tilde{z}$; detection replaces the non-transmitted layers with zero tensors and combines them with $\tilde{z}$ via an FPN~\cite{lin2017fpn}. The backbone $\theta$ is frozen; only the CAM and task heads are updated.

\subsection{Feature Compressor, Channel, and Feature Restorer}
Transmitting $z$ directly faces two challenges: insufficient channel capacity for the full $C$-dimensional feature, and noise corruption of discriminative structure. The CAM addresses both:
\begin{equation}
\tilde{z} = D_{\psi}\!\left(\mathcal{H}_{\gamma}(E_{\phi}(z))\right),
\end{equation}
where $E_{\phi}$ is the feature compressor ($C \to C'$, $C' \ll C$), $\mathcal{H}_{\gamma}$ denotes the channel at SNR $\gamma$, and $D_{\psi}$ is the feature restorer ($C' \to C$). Each component is detailed below.

\subsubsection{Feature Compressor}
The feature compressor applies a learned $1\!\times\!1$ convolution with group normalization~\cite{wu2018groupnorm} to reduce dimensionality from $C$ to $C'$ ($C' \ll C$):
\begin{equation}
\bar{z}_{:,i,j} = \mathrm{GN}\!\left(\mathbf{W}_e\, z_{:,i,j} + \mathbf{b}_e\right),
\end{equation}
where $\mathbf{W}_e \in \mathbb{R}^{C' \times C}$ and $\mathbf{b}_e \in \mathbb{R}^{C'}$. The encoded $\bar{z} \in \mathbb{R}^{C' \times H' \times W'}$ reduces transmitted symbols per spatial position by $C/C'$, yielding overall compression ratio $\rho = C'H'W'/(3HW)$ combined with the backbone's spatial downsampling. Group normalization stabilizes the feature distribution before transmission.

\subsubsection{Channel Model}
Dynamic satellite downlink conditions are abstracted as an AWGN channel with random-SNR sampling, deferring propagation-specific effects (fading, Doppler) to future work. The channel output is $r = \bar{z} + n$, $n \sim \mathcal{N}(0, \sigma^{2}\mathbf{I})$, with $\gamma$ sampled from:
\begin{equation}
\gamma \sim \text{Uniform}(\Gamma), \quad \Gamma = \{20, 13, 10, 0, -10, -13, -20\}\ \text{dB}.
\end{equation}
The noise variance is set by the average signal power:
\begin{equation}
P_s = \frac{1}{C'H'W'} \|\bar{z}\|_2^2, \qquad \sigma^{2} = P_s \cdot 10^{-\gamma/10}.
\end{equation}
Noise is injected during both training and inference. Random-SNR training prevents overfitting to a single operating point.

\subsubsection{Feature Restorer}
The feature restorer applies a learned $1\!\times\!1$ convolution with group normalization to expand the noise-corrupted features back to $C$ dimensions:
\begin{equation}
\tilde{z}_{:,i,j} = \mathrm{GN}\!\left(\mathbf{W}_d\, r_{:,i,j} + \mathbf{b}_d\right),
\end{equation}
where $\mathbf{W}_d \in \mathbb{R}^{C \times C'}$ and $\mathbf{b}_d \in \mathbb{R}^{C}$. The recovered $\tilde{z} \in \mathbb{R}^{C \times H' \times W'}$ matches the original dimensions of $z$, so task heads require no architectural modification.

\subsection{Task Heads}
Each task head takes $\tilde{z}$ and produces task-specific predictions. For classification (CLS), GAP followed by a linear classifier yields:
\begin{equation}
\hat{p} = \text{Softmax}(\mathbf{W}_{c}\text{GAP}(\tilde{z}) + \mathbf{b}_{c}).
\end{equation}

For detection, zero-filled shallow layers and $\tilde{z}$ are fed into an FPN neck:
\begin{equation}
(\hat{s}, \hat{b}) = h_{\omega}^{(\text{HBB})}(\text{FPN}(\mathbf{0}_{1}, \ldots, \mathbf{0}_{L-1}, \tilde{z})), \quad \hat{b} = (x, y, w, h).
\end{equation}
For oriented detection:
\begin{equation}
(\hat{s}, \hat{b}^{o}) = h_{\omega}^{(\text{OBB})}(\text{FPN}(\mathbf{0}_{1}, \ldots, \mathbf{0}_{L-1}, \tilde{z})), \quad \hat{b}^{o} = (x, y, w, h, \alpha).
\end{equation}
For detection, only the recovered last-stage feature carries task information, while the shallower pyramid inputs are zero-filled placeholders required by the FPN interface. FPN and detection head parameters are jointly optimized with the CAM.

\subsection{Training Objective}
The CAM and active task head are jointly optimized with
\begin{equation}
\mathcal{L} = \mathcal{L}_{\text{task}}^{(t)} + \lambda \mathcal{L}_{\text{rec}},
\end{equation}
where $\lambda$ is a fixed weighting coefficient (default $\lambda=0.2$). The task losses are
\begin{align}
\mathcal{L}_{\text{task}}^{(\text{CLS})} &= \text{CE}(\hat{p}, y), \\
\mathcal{L}_{\text{task}}^{(\text{HBB})} &= \mathcal{L}_{\text{cls}}(\hat{s}, s) + \mathcal{L}_{\text{box}}(\hat{b}, b), \\
\mathcal{L}_{\text{task}}^{(\text{OBB})} &= \mathcal{L}_{\text{cls}}(\hat{s}, s) + \mathcal{L}_{\text{obox}}(\hat{b}^{o}, b^{o}).
\end{align}
The feature reconstruction loss enforces consistency between $z$ and $\tilde{z}$:
\begin{equation}
\mathcal{L}_{\text{rec}} = \frac{1}{CH'W'} \|\tilde{z} - z\|_{2}^{2}.
\end{equation}
This balances task effectiveness and feature-level recoverability. The backbone $\theta$ is frozen; only $\phi=\{\mathbf{W}_e, \mathbf{b}_e\}$, $\psi=\{\mathbf{W}_d, \mathbf{b}_d\}$, and task head $\omega$ are updated.
\begin{figure*}[t]
    \centering
    \includegraphics[width=\textwidth]{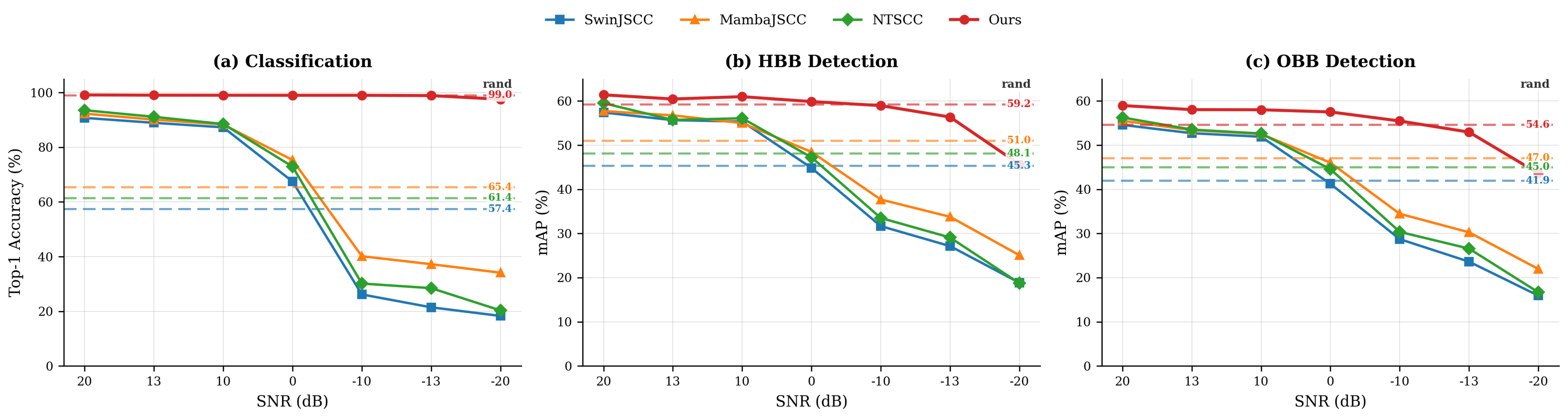}
    \caption{Performance across SNR conditions for (a)~classification on EuroSAT (Acc., $\rho=1/96$), (b)~HBB detection on DIOR (mAP, $\rho=1/15$), and (c)~OBB detection on DIOR-R (mAP, $\rho=1/15$). Dashed lines: random-SNR performance. The proposed method outperforms all compared methods, with the gap widening at low SNR.}
    \label{fig:comparison}
\end{figure*}

\subsubsection{Task-Specific Implementation Details}
\section{Experiments}

\subsection{Experimental Details}

This subsection describes the pretraining dataset, downstream evaluation datasets, the common evaluation protocol, and task-specific implementation details used throughout the experiments.

\subsubsection{Pretraining Dataset}
SAMRS~\cite{wang2023samrs} provides joint supervision for semantic segmentation, instance segmentation, and rotated object detection. It comprises three subsets (SOTA, SIOR, FAST) derived from DOTA-V2.0~\cite{xia2018dota}, DIOR~\cite{li2020dior}, and FAIR1M-2.0~\cite{sun2022fair1m} by converting rotated bounding-box annotations with SAM~\cite{kirillov2023sam}, totaling 105,090 images and 1,668,241 instances at canonical sizes $1024\!\times\!1024$, $800\!\times\!800$, and $600\!\times\!600$. For each rotated box, SAM produces a binary instance mask; the minimum enclosing horizontal rectangle serves as the axis-aligned box, and semantic maps are formed by assigning each box's category to its mask pixels.

\subsubsection{Downstream Datasets}
\textbf{EuroSAT}~\cite{helber2019eurosat} contains 27,000 Sentinel-2 images ($64\!\times\!64$) across 10 land-use classes. We use the public train/val split.
\textbf{DIOR}~\cite{li2020dior} contains 23,463 images ($800\!\times\!800$) across 20 categories with 192,472 HBB instances, split into 5,862/5,863/11,738 for train/val/test. We report results on the test set.
\textbf{DIOR-R}~\cite{li2020dior} provides OBB annotations for the same 23,463 DIOR images (192,158 instances, parameterized by center, width, height, angle). The same split is used.

\subsubsection{Evaluation Protocol}
All methods use the same frozen IMP+MTP backbone and the same downstream head architecture. For each method, the downstream head is trained under the same protocol on top of the corresponding communication pipeline. The proposed method extracts $z$ at the transmitter, transmits it through the CAM, and feeds $\tilde{z}$ directly into the task head. The compared methods (SwinJSCC~\cite{yang2025swinjscc}, MambaJSCC~\cite{wu2026mambajscc}, NTSCC~\cite{dai2022ntscc}) are trained on the same split under the same compression ratio and random-SNR channel using their original encoder--decoder architectures; at test time, each method reconstructs an image that is passed through the same frozen backbone and task head. Task heads are trained identically atop the frozen backbone, and evaluation uses identical SNR conditions. This isolates a single factor: whether the communication layer transmits task-oriented features (proposed) or reconstructed images (compared methods). An AWGN channel is adopted, with the model trained under randomly sampled SNR from $\Gamma = \{20, 13, 10, 0, -10, -13, -20\}$~dB.

\paragraph{Classification (EuroSAT)}
Images are resized to $224\!\times\!224$. The backbone produces four-stage features with channels $(192, 384, 768, 1536)$ and spatial sizes $(56, 28, 14, 7)$. Only $z \in \mathbb{R}^{1536 \times 7 \times 7}$ is transmitted, compressed to $C'=32$ ($\rho \approx 1/96$). The classification head applies GAP + linear classifier on the recovered $1536$-d feature. Training uses AdamW (lr $2\!\times\!10^{-5}$, weight decay $0.05$) with 5-epoch linear warmup from $10^{-6}$ followed by cosine annealing, batch size $64$, $100$ epochs. Augmentation: random resized crop, flip, RandAugment, random erasing. Loss weight $\lambda = 0.2$. Metric: Top-1 classification accuracy (Acc., \%).

\paragraph{HBB Detection (DIOR)}
The backbone produces four-stage spatial resolutions $(200, 100, 50, 25)$. The payload $z \in \mathbb{R}^{1536 \times 25 \times 25}$ is compressed to $C'=205$ ($\rho \approx 1/15$). The recovered $\tilde{z}$ is combined with zero-filled shallow layers and fed into an FPN~\cite{lin2017fpn} (256 channels) + Faster R-CNN~\cite{ren2017fasterrcnn}. Training uses AdamW (lr $1\!\times\!10^{-4}$, weight decay $0.05$) with 5-epoch warmup from $10^{-6}$, cosine annealing, batch size $8$, $36$ epochs. Augmentation: horizontal flip, resized crop, multi-scale training (shorter side $\in [640, 800]$). $\lambda = 0.2$. Metric: mean average precision (mAP, \%).

\paragraph{OBB Detection (DIOR-R)}
All settings match the HBB configuration except that the detection head is Oriented R-CNN~\cite{xie2021orientedrcnn} and random rotation is added to augmentation. Metric: mAP(\%).

\begin{figure*}[t]
    \centering
    \includegraphics[width=\textwidth]{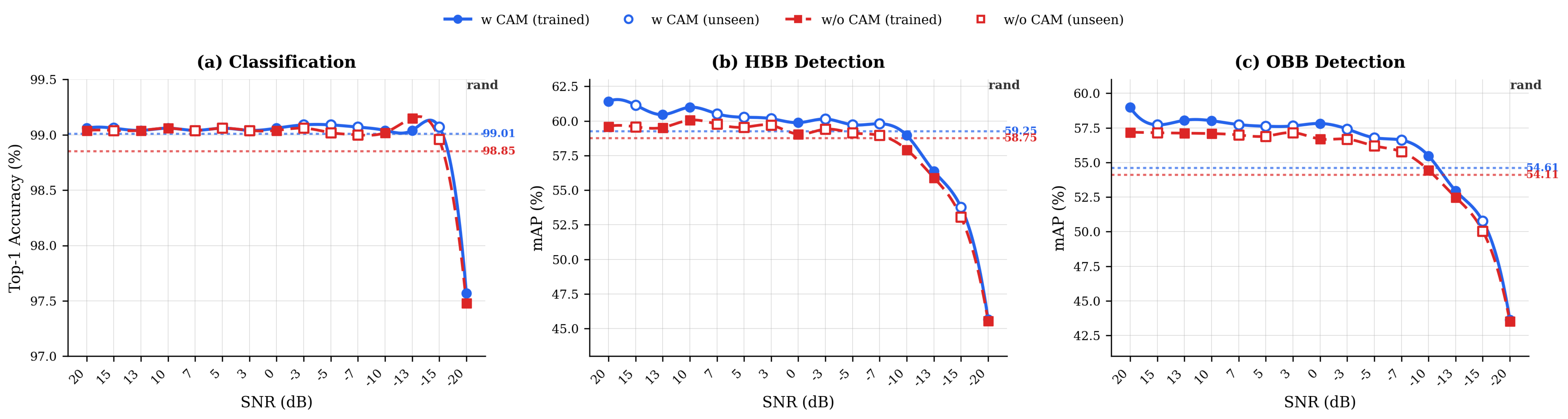}
    \caption{Ablation of CAM across SNR conditions for (a) classification, (b) HBB detection, and (c) OBB detection. Filled markers denote trained SNR points, hollow markers denote unseen points, and dashed lines denote random-SNR performance.}
    \label{fig:ablation}
\end{figure*}
\subsection{Comparison with Reconstruction-Oriented SOTA Methods}

We compare the proposed method against three reconstruction-oriented SOTA methods, SwinJSCC~\cite{yang2025swinjscc}, MambaJSCC~\cite{wu2026mambajscc}, and NTSCC~\cite{dai2022ntscc}, on all three downstream tasks under the evaluation protocol described above. Performance is reported at each SNR in $\Gamma$ and under a random condition $\gamma \sim \text{Uniform}(\Gamma)$. This comparison directly tests whether transmitting task-oriented features is more effective than transmitting images that must later be reconstructed for inference.

\subsubsection{Scene Classification}
Fig.~\ref{fig:comparison}(a) summarizes classification accuracy under all trained SNR conditions.

Under the random condition, the proposed method achieves $99.01$\% versus $65.37$\% for MambaJSCC. At $-20$~dB the gap widens to $97.57$\% versus $34.12$\%. Even at $20$~dB, feature transmission leads ($99.15$\% versus $93.56$\% for NTSCC), showing that task-oriented features are a better communication target even under favorable conditions.

The proposed method maintains $\geq 97.57$\% across the entire SNR range, while compared methods drop to $18$--$34$\% at $-20$~dB. This stability is likely due to the fact that classification depends primarily on high-level semantics in the last-stage features, while GAP reduces sensitivity to local perturbations. Notably, this behavior indicates that preserving semantic separability is more important than preserving pixel fidelity for this task.

\subsubsection{Horizontal Object Detection}
Fig.~\ref{fig:comparison}(b) summarizes horizontal detection mAP under all trained SNR conditions.

Under the random condition, the proposed method achieves $59.25$\% versus $51.01$\% for MambaJSCC. The gap grows as SNR decreases: $+1.89$ at $20$~dB, $+11.41$ at $0$~dB, $+21.27$ at $-10$~dB, $+20.58$ at $-20$~dB. Detection is more sensitive to spatial structure than classification; reconstruction artifacts can disrupt localization cues, whereas feature-domain transmission avoids an explicit image reconstruction stage and therefore tends to preserve task-relevant structure more directly.

\subsubsection{Rotated Object Detection}
Fig.~\ref{fig:comparison}(c) summarizes oriented detection mAP under all trained SNR conditions.

Under the random condition, the proposed method achieves $54.61$\% versus $47.05$\% for MambaJSCC. At $-20$~dB the gap widens to $43.63$\% versus $21.97$\%, and at $-10$~dB to $55.48$\% versus $34.46$\%. Compared methods degrade more steeply than in HBB detection, as oriented boxes are sensitive to local geometry and boundary precision. Furthermore, angle prediction depends on fine-grained structural consistency, which is particularly vulnerable to reconstruction noise.

Across all three tasks, the proposed method's drop from $20$ to $-20$~dB is $\Delta_{\text{CLS}} = -1.58$ Acc., $\Delta_{\text{HBB}} = -15.74$ mAP, $\Delta_{\text{OBB}} = -15.35$ mAP. Detection is far more SNR-sensitive than classification, as it relies on spatially precise localization cues vulnerable to noise. These cross-task differences further support the need to align the transmitted representation with the actual downstream objective.

\subsection{Ablation Study}

To verify the effectiveness of the feature restorer in the CAM, we compare two variants: \textbf{w/o CAM} (compressor only) and \textbf{w CAM} (full CAM with restorer) in Fig.~\ref{fig:ablation}.

\subsubsection{Experimental Setup}
The \textbf{w/o CAM} variant compresses from $C=1536$ to $C'$ ($C'=32$ for classification, $C'=205$ for detection) and injects AWGN noise, but removes the restorer. The task head is adapted to accept $C'$-dimensional input directly. The \textbf{w CAM} variant retains the full CAM: after compression and noise injection, the restorer expands features back to $C$ dimensions. Both variants share the same frozen backbone and random-SNR training protocol.

\subsubsection{Results and Analysis}
The w CAM variant consistently outperforms w/o CAM. Under the random condition, the restorer improves HBB mAP by $+0.50$ (59.25 vs.\ 58.75) and OBB by $+0.50$ (54.61 vs.\ 54.11). Classification improves by $+0.16$ Acc.\ (99.01 vs.\ 98.85), as expected since GAP-based classification is less sensitive to dimensionality than FPN-based detection. The advantage is consistent across SNR: $+1.03$ at $-10$~dB for both HBB and OBB, $+1.79$ at $20$~dB where dimensionality expansion helps the FPN produce richer features, and $+0.10$ at $-20$~dB where the noise floor dominates. The compressor performs the critical compression, while the restorer provides consistent refinement for detection.

\subsection{Complexity Analysis}

This subsection compares the parameters, FLOPs, and latency of transmission-related modules only; the shared backbone, FPN, and task heads are excluded. FLOPs for methods with custom CUDA operators may be partially estimated; wall-clock latency is measured with \texttt{torch.cuda.Event}. 

\begin{table}[t]
\centering
\caption{Complexity of transmission-related modules under the detection setting (DIOR-R, $C'=205$, feature $1536 \times 25 \times 25$). Backbone, FPN, and task heads are excluded. $^\dagger$FLOPs partially estimated due to custom CUDA operators.}
\label{tab:complexity}
\begin{threeparttable}
\begin{tabular}{lccc}
\toprule
Method & Params & FLOPs & Latency (ms) \\
\midrule
Ours (compressor only) & 315.29K & 197.44M & $0.12 \pm 0.00$ \\
Ours (full CAM) & 633.24K & 399.04M & $0.47 \pm 0.19$ \\
\midrule
SwinJSCC & 33.07M & 34.02G & $18.23 \pm 1.05$ \\
MambaJSCC & 14.55M & 6.51G$^\dagger$ & $16.52 \pm 0.94$ \\
NTSCC & 28.63M & 1.76G & $4.47 \pm 0.28$ \\
\bottomrule
\end{tabular}
\begin{tablenotes}[flushleft]
\footnotesize
\item \textit{Note}: Latency measured on one GPU with 200 iterations.
\end{tablenotes}
\end{threeparttable}
\end{table}

Table~\ref{tab:complexity} compares modules under the detection setting ($C'=205$, feature $1536\!\times\!25\!\times\!25$). The CAM uses $633.24$K parameters and $399.04$M FLOPs—$1$--$2$ orders of magnitude below the compared methods ($14.55$--$33.07$M parameters, $1.76$--$34.02$G FLOPs). Latency is $0.47$~ms versus $4.47$--$18.23$~ms for the compared methods. Under the classification setting ($C'=32$, feature $1536\!\times\!7\!\times\!7$), the overhead is even smaller: $101.44$K parameters, $5.20$M FLOPs, and $0.22$~ms latency. This efficiency results from using only two $1\!\times\!1$ convolutions on compact backbone features rather than full-resolution images.

\section{Conclusion}
This paper presents a task-oriented semantic feature transmission framework for multi-task satellite remote sensing over varying-SNR channels. Instead of reconstructing images before inference, the proposed framework directly transmits task-relevant semantic features extracted by a multitask-pretrained backbone and performs downstream inference on recovered features. The lightweight channel adaptation module preserves discriminative feature quality under channel noise, leading to strong classification and detection performance with smooth degradation across the trained SNR range. Under the same compression ratios and AWGN settings, the framework consistently outperforms reconstruction-oriented JSCC methods, with the most pronounced gains in the low-SNR regime.

Future work will extend the channel model beyond AWGN to fading, Doppler, and time-selective channels representative of LEO satellite links, incorporate adaptive rate control with variable compression ratios and explicit payload accounting, and explore progressive multi-scale semantic feature transmission for small-object localization and oriented detection. Hardware-aware deployment with lightweight onboard backbones, model compression, and energy-aware inference on resource-constrained satellite platforms will also be investigated for practical implementation.

\section{Acknowledgement}
This work has been funded by the National Natural Science Foundation of China under Grant 62301063.

\bibliographystyle{IEEEtran}
\bibliography{references}

\end{document}